\documentstyle[aclap]{article}

\title{Towards a PURE Spoken Dialogue System for Information Access}
\author{Rajeev Agarwal\\
Media Technologies Laboratory\\
Texas Instruments Inc.\\
PO Box 655303, MS 8374 
Dallas, TX 75265\\
USA\\
rajeev@csc.ti.com}

\input{psfig}

\begin{document}

\maketitle

\vspace{0.1in}
\begin{abstract}

With the rapid explosion of the World Wide Web, it is becoming increasingly
possible to easily acquire a wide variety of information such as flight
schedules, yellow pages, used car prices, current stock prices, entertainment
event schedules, account balances, etc.  It would be very useful to have
spoken dialogue interfaces for such information access tasks.  We identify
portability, usability, robustness, and extensibility as the four primary
design objectives for such systems.  In other words, the objective is to
develop a PURE (Portable, Usable, Robust, Extensible) system.  A two-layered
dialogue architecture for spoken dialogue systems is presented where the upper
layer is domain-independent and the lower layer is domain-specific.  We are
implementing this architecture in a mixed-initiative system that accesses
flight arrival/departure information from the World Wide Web.

\end{abstract}

\section{Introduction}

With the rapid rate at which the availability of information is increasing, it
is important to make access to this information easier.  One may wish to get
the arrival/departure information for a given flight, verify if a particular
book is available at a library, find the stock price for any fund, access
yellow page information on-line, check/maintain voice mail remotely, get
schedules for entertainment events, perform remote banking transactions, get
used car prices, and the list goes on and on.  Such tasks can be classified as
{\em information access} (IA) tasks, where the primary objective is to get
some piece of information from a certain place by providing constraints for
the search.  Some of these tasks may also involve an ``action'' that may
change the state of the underlying database, e.g., making a reservation for an
event, making transactions on an account, etc.  It would be very helpful to
develop Spoken Dialogue (SD) interfaces for such IA applications, and several
such attempts are already being made
\cite{SenGodPao-96,SadFerCoz-96,AbeBroBun-96,FraDal-96,LamGauBen-96,KelRueSei-96,Nie-96,BarSin-96,GorParSac-96}.

In this paper, we differentiate between such IA tasks and the more complicated
{\em problem solving} tasks where multiple sub-problems are concurrently
active, each with different constraints on them and the final solution
consists of identifying and meeting the user's goals while satisfying these
multiple constraints.  Examples of such applications include a system that
offers investment advice to a user based on personal preferences and the
existing market conditions, or an ATIS-like application that assists the user
in travel planning including flight reservations, car rental, hotel
accommodations, etc.

\begin{figure*}[t]
  \begin{center}
    \leavevmode \psfig{figure=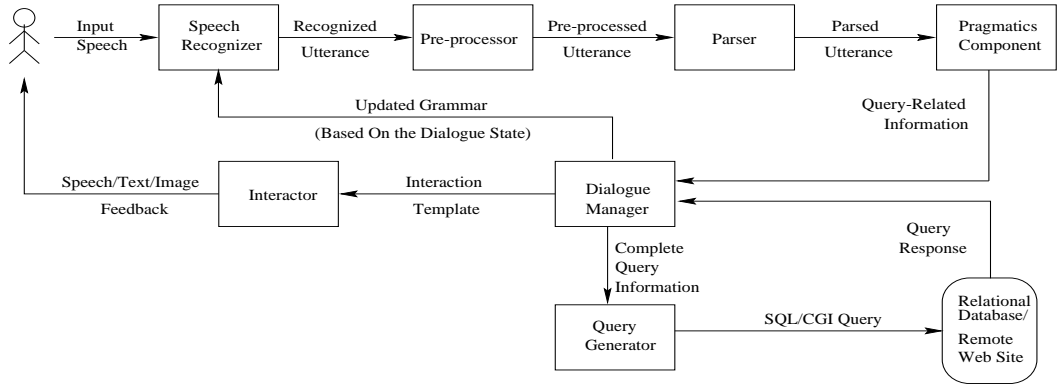,width=5.5in,height=2in}
  \end{center}
  \caption{Outline of the Components of the Spoken Dialogue System}
  \label{fig:vq}
\end{figure*}

In addition to the general requirement of accuracy, there are four other
important design objectives for SD systems:

\begin{itemize}

\item Portability of an SD system refers to the ability of the system to be
	moved from one application/domain to another. 

\item Usability of an SD system refers to the ease with which a user can use
	the system and the naturalness that it provides.

\item Robustness of an SD system refers to the ability of the system to help
	the user acquire the desired information even in the presence of user
	and system errors. 

\item Extensibility of an SD system implies that additional queries within a
	given application can be added to the system without much trouble.

\end{itemize}

The purpose of this paper is to describe an SD system, in particular the
dialogue manager, that is being developed with these objectives in mind.
Since these design objectives are often conflicting in nature, one has to
strike a balance between them.  In a manner of speaking, one could say that
the objective is to create a PURE (Portable, Usable, Robust, Extensible)
system.  It is our belief that it is possible to develop an ``almost'' PURE
system for IA tasks.

\section{Overall System Description}

The overall SD system is responsible for taking user utterances as input,
processing them in a given context in an attempt to understand the user's
query, and satisfying his/her request.  The user does not need to know
anything about the structure of the database or the architecture of the
system.  In case the user's utterance has missing, ambiguous, inconsistent, or
erroneous information, the system engages the user in a dialogue to resolve
these.  The system is designed to be mixed-initiative, i.e., either the user
or the system can initiate a dialogue or sub-dialogue at any time.  The
dialogue ends when the user decides to quit the system.  The system can be
used for querying a relational database using SQL or invoking a
CGI\footnote{CGI stands for Common Gateway Interface.  It is a tool that
assists web programmers in creating interactive, user-driven applications.
Several web sites permit database queries where the user types in the search
constraints on an HTML FORM and the server submits this form to the CGI script
which generates a response after searching a local database.  Note that here
we refer to such database searches and not to the string searches as offered
by Lycos, WebCrawler, Excite, etc.} script on the web.  A brief overview of
the different components is presented in Figure~\ref{fig:vq}.

\begin{itemize}

\item Speech Recognizer: It is responsible for recognizing the user utterance
	and producing a recognition string.  We currently write separate
	context-free grammars for each state of the dialogue and use these to
	recognize the utterances with the DAGGER speech recognition system
	described in \cite{HemThr-95}.  An important feature of this
	recognizer is that based on the dialogue state, certain grammars may
	be switched into or out of the dynamic vocabulary\footnote{We only use
	the grammar switching feature of DAGGER, but it offers the ability
	to load completely new grammars dynamically if such a need arises.},
	thereby leading to better speech recognition accuracy.

\item Preprocessor: This component is responsible for identifying
	domain-independent (e.g., time, place name, date) and
	domain-specific semantic patterns (e.g., airport name, book title)
	in the input utterance.

\item Parser: Since user utterances could be ungrammatical in nature, a
	partial parser has been implemented to parse the input utterance into
	its component phrases.  This provides added robustness, although lack
	of a deep structure in the parse sometimes causes the pragmatics
	component to miss useful information.

\item Pragmatics Component: This component is responsible for identifying the
	values of relevant fields that are specified in the utterance, based
	on the partial parse of the utterance.  It uses an application
	specific input file called the {\em application schema}, which
	describes all the relevant fields in that application and
	lexico-semantic patterns that indicate their presence.  It also
	describes the possible queries that may be made in that application.

\item Dialogue Manager: It evaluates the knowledge extracted by the pragmatics
	component to determine the current state of the dialogue.  It
	processes this new dialogue state and constructs an ``interaction
	template'' that determines what feedback should be provided to the
	user.

\item Query Generator: This component is responsible for generating a database
	query.  It can generate either a SQL query for a relational database
	or a CGI script query for querying a web site.
	
\item Interactor: It is responsible for converting the interaction template
	generated by the dialogue manager into English sentences that can be
	printed and/or spoken (using a text-to-speech system) to the user to
	provide feedback.  It uses a template-to-string rules file that
	contains rules for all possible types of interactions.  In some cases,
	it may also provide feedback by updating a displayed image.

\end{itemize}

This gives a brief overview of our SD system.  The system is still under
development, and is being tested on the flight arrival/departure information
application for which we query the American Airlines web site \cite{AA}.
System development is expected to be completed soon.  We have also used this
system to begin developing a ``Map Finder'' demo that queries the MapQuest web
site \cite{MQ} to display maps of any street address or intersection in the
United States.  We intend to port this system to the yellow pages information
access application in the near future.

\section{Dialogue Manager Design}

\subsection{Background}

Existing approaches to designing dialogue managers can be broadly classified
into three types: graph-based, frame-based, and plan-based.  This section
gives a brief overview of these approaches and argues that for IA tasks, the
frame-based approaches are the most suitable.

{\em Graph-based} approaches require the entire dialogue state transition
graph for an application to be pre-specified.  Several dialogue design
toolkits are available to assist developers in this task, such as the SLUrp
toolkit \cite{SutNovCol-96}, SpeechWorks toolkit \cite{ALT}, or DDL-tool
\cite{Bae-96}.  It is often cumbersome and sometimes impossible to pre-specify
such a dialogue graph.  Further, such approaches are not robust as they cannot
appropriately handle any unforeseen circumstances.

{\em Plan-based} approaches attempt to recognize the intentions of the
entities involved in the discourse and interpret future utterances in this
light.  They are usually based on some underlying discourse model, several of
which have been developed over the years
\cite{CohPer-79,ManTho-83,GroSid-86,Car-90}.  We argue here that although
plan-based systems are very useful for problem-solving tasks like the ones
described earlier, that degree of sophistication is not needed for IA tasks.
For example, of the five types of intentions outlined by Grosz and Sidner
(1986), only ``intent that some agent believe some fact'' and ``intent that
some agent know some property of an object'' are encountered in IA tasks, and
they can be easily conflated for such tasks, without any loss of information.
Further, although modeling a speaker's intentions and the relations between
them is informative about the structure of the discourse, their recognition in
an actual system may be non-trivial and prone to errors.  Most IA tasks have
only one discourse purpose, and that is to get some information from the
system.  The various discourse segments are all directed at providing the
system with relevant constraints for the database query.  Therefore, explicit
modeling of the discourse purpose or discourse segment purpose is unnecessary.

{\em Frame-based} systems typically have a domain/application model to which
they map user utterances in an attempt to recognize the nature of the user's
query.  The constraints of the application drive the analysis of utterances.
Such systems usually ignore phenomena like diectic references, expressions of
surprise, discourse segment shifts, etc.

\begin{figure*}[t]
  \begin{center}
    \leavevmode \psfig{figure=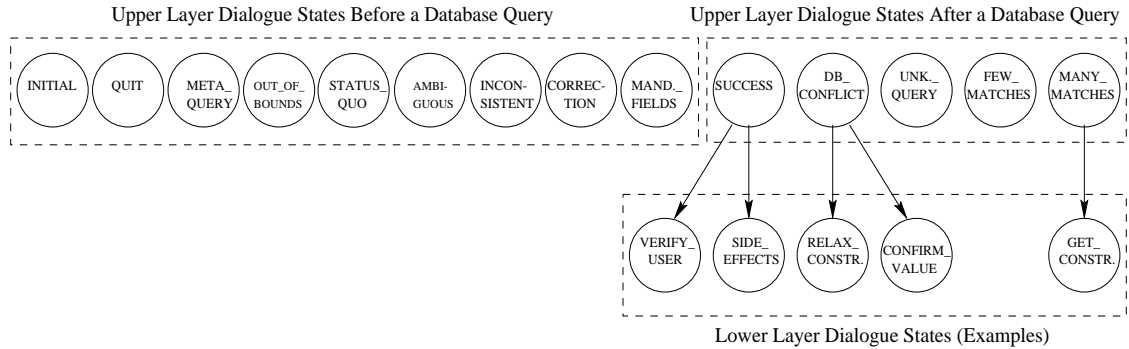,width=6in,height=1.8in}
  \end{center}
  \caption{States in the Two-Layered Dialogue Management Architecture}
  \label{fig:dm}
\end{figure*}

\subsection{Two-Layered Architecture}

It is our contention that for IA tasks, the dialogue between the user and the
system proceeds in a domain-independent manner at a higher level and can be
described by a set of domain-independent states.  Some domain-specific
interactions are required once the dialogue is in one of these higher level
states and these can be described by a different set of states.  This view of
the structure of the dialogue led us to a two-layered architecture for the DM.
The upper layer is completely domain-independent, while the lower layer has
dialogue states that constitute domain-specific sub-dialogues.  Further,
although the different states of the dialogue are pre-specified, the system
automatically identifies what state it is in based on the user's utterance,
the result of the database query, and knowledge of the previous dialogue state.
This is what Fraser and Dalsgaard (1996) refer to as a {\em self-organizing}
system.  Most plan-based and frame-based systems are self-organizing.  The
states in the DM are shown in Figure~\ref{fig:dm} and are described in detail
in this section.

\subsubsection{Dialogue States}

All fourteen states presented here at the top level belong to the upper layer
of the dialogue.  For some of these upper layer states, references are made to
the lower layer states that they may spawn to accomplish domain-specific
sub-dialogues.  After every user utterance, the DM checks to see if the
dialogue is in one of the upper layer dialogue states.  Lower layer states are
checked only if the system is already in a sub-dialogue.  The upper layer
states are tried in the order in which they are described below since if the
dialogue is in any of the earlier states, there is no point in trying later
ones.  The existence of one of the first nine states listed below may be
determined without a database query.  If the dialogue is not in any one of
these nine states, then there is enough information to issue a query, and the
dialogue may be in one of the last five states based on the results of the
query.  The dialogue ends when the QUIT state is reached.

\begin{enumerate}

\item {\em INITIAL}: This is the state in which each dialogue starts and
reverts to after a query made by the user has been completely processed.

\item {\em QUIT}: If the system detects that the user wants to
terminate the current dialogue, then the dialogue enters this state. 

\item {\em META\verb$_$QUERY}: The dialogue reaches this state when the user
either explicitly asks for help (e.g., ``Please help me,'' ``what can
I say,'' etc.) or asks for some meta-level information about the system's
capabilities (e.g., ``what cities do you know about?'').  The help messages in
the system are context-sensitive and are based on the current dialogue state.

\item {\em OUT\verb$_$OF\verb$_$BOUNDS}: This state is reached when the system
realizes that the user either wants to access information that the system is
not equipped to handle or access ``legitimate'' information in ways the system
is not designed to handle.  For example, if a system is designed to access
American Airlines flight information and the user says ``what time does Delta
flight 472 reach Dallas?,'' the system enters the OUT\verb$_$OF\verb$_$BOUNDS
state.  An example of an improper legitimate query could be ``what time does
my plane leave?,'' if the system expects the word 'flight' but not 'plane'.
The objective is not just to quit gracefully, but to allow the user to
re-enter the dialogue at some place.  In the first case, the system informs
the user of the limitations of the system, switches the dialogue to the
INITIAL state, and permits the user to revert to some query within the bounds
of the system.  In the second case, it informs the user that the word 'plane'
is unknown to the system, and requests him/her to rephrase the query.

\item {\em STATUS\verb$_$QUO}: This state is reached if the system determines
that the most recent utterance by the user provided no additional
query-related information to the system.  This is an indication that the user
was either completely silent, did not know the answer to the system's previous
question (may have responded by saying ``I don't know'' to something the
system had asked), explicitly asked the system to repeat the last feedback
(may have said ``Can you repeat that''), the speech recognizer misrecognized
the part of the utterance that was meant to be informational, or the utterance
really had no new information.  Based on what the user said, an appropriate
response is generated.

\item {\em AMBIGUOUS}: This state is reached when one of three types of
ambiguities exists in the system.  {\em Lexical ambiguity} arises if some user
term matches two entities within the same semantic class.  For example, in a
library application, if the user asks for ``Dickens'' and the database
contains two or more authors with that last name, this term is lexically
ambiguous.  {\em Class ambiguity} arises if a term may belong to two or more
semantic classes.  In the above example, if there is also a book entitled
``Dickens'' in the database, then class ambiguity exists since it is unknown
whether the user meant the 'author' or the 'title'.  This can often be
resolved based on the surrounding context.  {\em Field ambiguity} arises when
the system has found a term that could refer to more than one database field.
For example, in a flight arrival/departure application, if the system prompts
the user for either the arrival city or departure city, and the user just says
``Newark,'' the field to which the term belongs is ambiguous.

\item {\em INCONSISTENT}: User or system errors may sometimes lead the DM to
this state where the system's knowledge of the various fields violates some
consistency rule.  The consistency rules specific to an application are
provided in an input file.  For example, an error may cause the system to
believe that the departure city and the arrival city in a flights
arrival/departure application are the same.  If that happens, the user is
notified of the inconsistency so that the error may be rectified.

\item {\em CORRECTION}: This state is reached when the system realizes that
the user is attempting to correct either an error the user may have made or an
error made by the recognizer.  As a result, the system accepts the corrected
value provided by the user (assuming that this new value is correctly
recognized) and provides appropriate feedback.  For example, in a flight
arrival/departure application, the user might say ``I said Dallas, not
Dulles'' to correct a misrecognition by the speech recognizer.

\item {\em MANDATORY\verb$_$FIELDS}: This state is needed only for
applications in which values for certain fields must be known before a query
can be issued.  This is often true of applications that invoke CGI scripts on
the web.  For example, the American Airlines web site only permits a query if
the user specifies either the flight number, or the arrival and departure city
and approximate arrival time, or the arrival and departure city and
approximate departure time.  This state ensures that values for these
mandatory fields are obtained from the user before issuing a CGI query.
 
\item {\em SUCCESS}: If none of the previous states were found, a query is
issued to the system.  If this query results in a successful match, then the
dialogue is in this state.  After providing appropriate feedback to the user,
the system performs a further check to see if any ``action'' needs to be
carried out on the accessed item(s) of information.  For example, in a banking
application, having checked the balance in a savings account, the user may now
wish to transfer money from checking to savings.  This state usually spawns a
sub-dialogue which may or may not be domain-specific.  The lower level
dialogue states in this sub-dialogue could be -

\begin{itemize}

\item {\em VERIFY\verb$_$USER}: which asks for the user's account ID and password,

\item {\em SIDE\verb$_$EFFECTS}: which informs the user of some side effects
	of the imposed constraints, e.g. ``This transaction will lead to a
	negative balance in the checking account,'' or

\item some other domain-specific state depending upon the nature of the action
	involved.

\end{itemize}

Once in this state, the user may start a new query, ask for more information
about the matched item, or quit the system.

\item {\em DATABASE\verb$_$CONFLICT}: A database conflict arises when the
constraints specified by the user do not match any item in the database.  This
could be because of conflicting information from the user or speech
recognition errors.  Such conflicts must be resolved before proceeding in the
dialogue.  Conflict resolution may be accomplished by a sub-dialogue in the
lower layer.  Some of the possible states in the lower layer are:

\begin{itemize}

\item {\em RELAX\verb$_$CONSTRAINT}: asks the user to relax a certain
	constraint, e.g., ``No Thai restaurant found on Legacy, but there is
	one on Spring Creek -- is that OK?'' (the system needs domain-specific
	information that Legacy and Spring Creek are close to each other).  In
	some cases, the system also needs to know which constraints are
	``negotiable''.

\item {\em CONFIRM\verb$_$VALUE}: asks the user to confirm some field values
	provided by the user.  The confirmation is needed to ensure that it
	was not a system or user error that caused a conflict.

\end{itemize}

\item {\em UNKNOWN\verb$_$QUERY}: In most applications, the user may query for
different types of information.  In a yellow pages application, for example,
the user may ask about a phone number, an email address, or a postal address.
The DM may need to know what item of information the user is interested in, as
this determines the feedback provided to the user.  This is especially useful
in applications without a display (queries made over the telephone) since it
takes time to give more information than is necessary.  Note that it is often
possible to issue a database query even if this information is not known, and
that is why this state belongs to the set of possible states after a query has
been made.

\item {\em FEW\verb$_$MATCHES}: If the database query results in a ``few''
matches, then the dialogue enters this state.  Whenever few matches are found,
the most efficient way to consummate the query is to enumerate these matches
so the user can the select the one of interest.

\item {\em MANY\verb$_$MATCHES}: If none of the previous states are reached,
the database query must have resulted in too many matches, i.e., not enough
information was supplied by the user to match only a single or a few database
items.  This state may spawn a domain-specific sub-dialogue in the lower
layer, one of whose states could be:

\begin{itemize}

\item {\em GET\verb$_$CONSTRAINT}: The objective is to ask the user to specify the
	least number of constraints that lead to the SUCCESS state.  So, whenever
	possible, this dialogue state identifies what piece of information
	would be ``most informative'' at that point in time, and asks the user
	to specify its value.

\end{itemize}

\end{enumerate}

This concludes the description of the various dialogue states.  While we have
attempted to provide an upper layer that covers most IA tasks, the lower layer
states given here are just examples of some possible states.  Depending upon
the application, more lower layer states can be added to improve the
usability/robustness of the system.

\section{Comparison to Other Approaches}

Several other mixed-initiative spoken dialogue systems have been developed for
information access tasks
\cite{AbeBroBun-96,BenDevRos-96,KelRueSei-96,SenGodPao-96,FraDal-96,SadFerCoz-96,BarSin-96}
and they provide varying degrees of dialogue management capability.  Our
dialogue management approach is possibly most similar to that proposed by
Abella et al. (1996), with some important differences.  We have attempted to
clearly define a comprehensive set of states to handle various contingencies
including out-of-bounds queries, meta-queries, ambiguities, and
inconsistencies due to user/system errors.  We feel that our two-layered
architecture should make the system more portable.  We further contend that if
one encounters a dialogue state that is not covered by our state set, it can
be abstracted to an upper level state which may later be useful in other
applications.  Abella et al.\ (1996) do present a nice question selection
methodology that we lack\footnote{It may be noted that such a methodology is
possible only with local relational databases.  It cannot be implemented when
querying CGI scripts on the web since we do not have access to the underlying
database.}.  We currently resort to a domain-dependent GET\verb$_$CONSTRAINT
state but hope to improve on that in the future.

The primary bottleneck in our system at this time is the parser which only
identifies partial parses and does not perform appropriate PP-attachment,
conjunct identification, or do anaphora resolution or ellipsis handling.  We
need to replace the existing partial parser with a better parser to improve
the overall system accuracy.

\section{How PURE is it?}

We started out by saying that the objective is to develop a PURE spoken
dialogue system for information access tasks.  We want to ensure that our
system aims to be as PURE as it can be.  In this section, we list those
features of our system that are intended to make it PURE.

\begin{itemize}

\item Portability:

\begin{itemize}
	
\item In order to move the SD system to a new domain, the following files must
	be specified: an {\em application schema} that was briefly described
	in Section~2; a {\em schema-to-database} mapping file that maps items
	in the application schema to the fields in the relational database or
	in the CGI script (e.g., the {\tt flight\verb$_$number} schema field
	maps to the {\tt fltNumber} field in the CGI script); a {\em
	user-to-database} mapping file that consists of the various ways a
	user may refer to a value of a database field (e.g., ``Big Apple''
	maps to ``New York''); and a {\em consistency-rules} file.

\item The two-layered architecture ensures that the overall dialogue
	progresses at a domain-independent level, and keeps the
	domain-independent and domain-specific states separate.

\item Self-organizing dialogue structure makes it more portable.

\item Partial parser can be directly ported to a new domain.

\end{itemize}

\item Usability:

\begin{itemize}

\item Mixed-initiative approach helps to promote usability.

\item Feedback provided by the interactor can be made more domain-friendly by
	specifying some extra domain-specific rules at the top of the {\em
	template-to-string} rules file, since these rules are executed in the
	order specified.

\item User may say ``I don't know,'' ``Please help me,'' ``What can I say,''
	etc.\ at any time to get some guidance.  The help messages are
	context-sensitive.

\item We intend to add prompt randomization, as suggested by Kellner et
	al. (1996) to make the interactions ``less boring.''

\item The OUT\verb$_$OF\verb$_$BOUNDS state and the META\verb$_$QUERY state
	improve usability by informing the user of why a certain utterance was
	inappropriate and allowing the user to ask about the system's
	abilities respectively.

\end{itemize}

\item Robustness:

\begin{itemize}

\item Partial parser can handle ungrammatical input.

\item Lexico-semantic pattern matching for field values ensures that
	misrecognition of a part of the utterance will still extract useful
	information from the correctly recognized part.

\item The CORRECTION and INCONSISTENT states increase the robustness of the
	system by making it possible to continue even in the presence of
	errors.

\end{itemize}

\item Extensibility:

\begin{itemize}

\item Additional queries can be added to any application by specifying the
	query semantics in the application schema and any new fields that
	they may need.

\end{itemize}

\end{itemize}

\section{Final Comments}

We have presented a dialogue management architecture that is mixed-initiative,
self-organizing, and has a two-layered state set whose upper layer is portable
to other applications.  The system is designed to generate either SQL queries
or CGI script queries, which makes it capable of querying the vast amount of
information available on the World Wide Web.  

Although the generation of CGI queries is driven by the schema-to-database and
user-to-database mappings files, some degree of application specific work
still needs to be performed.  One has to experiment with the web site and
study the source pages for the HTML FORMS screens in order to create these
mappings files and possibly write additional code to generate the appropriate
query.  For example, the American Airlines web site provides three different
web pages to support queries about flight arrival/departure information.  An
examination of all three source pages revealed that a hidden field {\tt
fltAns} gets one of three values based on which page invokes the script.  A
special hack had to be built into the query generator to assign an appropriate
value to this field.  Generation of proper user feedback requires us to also
examine the source page of the result of the query.  The main limitation of
querying CGI scripts is that if the web site being queried is modified by its
creators, slight modifications will have to be made to the query generator to
accommodate those changes.

Our initial experience with this system, especially porting it from the
flights arrival/departure application to the Map Finder application, has been
very encouraging.  Map Finder is a simpler task and some of the upper layer
states (UNKNOWN\verb$_$QUERY, FEW\verb$_$MATCHES, and MANY\verb$_$MATCHES)
never occur in this application.  An additional lower layer state called
{MAP\verb$_$COMMANDS} had to be implemented under the SUCCESS state to allow
the user to scroll the displayed map in any direction using spoken commands.
This required understanding the way the MapQuest web site handles these map
navigation commands.  The rest of the DM was easily ported to this new
application.

This system is still work-in-progress and more work remains.  We intend to
continue improving the existing components while also porting the system to
other applications so that we can learn from our porting experiences.

\section*{Acknowledgements}

The author wishes to thank Jack Godfrey for several useful discussions and his
comments on an earlier draft of this paper; Charles Hemphill for his comments
and for developing and providing the DAGGER speech recognizer; and the
anonymous reviewers for their valuable suggestions that helped improve the
final version of this paper.

\end{document}